\documentclass{article}
\usepackage{spconf,amsmath,amssymb,graphicx}

\usepackage{multirow}  
\usepackage{colortbl}  
\usepackage[table]{xcolor}  
\usepackage{booktabs}
\usepackage{tabularx}
\usepackage{amsmath}
\usepackage{subcaption}
\usepackage{hyperref}
\usepackage{soul}
\DeclareRobustCommand{\hlYellowOne}[1]{{\sethlcolor{table_best}\hl{#1}}}
\DeclareRobustCommand{\hlYellowTwo}[1]{{\sethlcolor{table_second}\hl{#1}}}

\definecolor{table_second}{HTML}{FFEDB3}
\definecolor{table_best}{HTML}{FFD44D}
\definecolor{table_cam_best}{HTML}{B0E0E6}
\definecolor{table_cam_second}{HTML}{E0FFFF}

\newlength\savewidth
\usepackage{caption}    
\usepackage{tabularx}
\usepackage{graphicx}
\usepackage{wrapfig}

\title{Pose-Free 3D Gaussian Splatting via Shape-Ray Estimation}
\name{Youngju Na\textsuperscript{*}, Taeyeon Kim\textsuperscript{*}, Jumin Lee, Kyu Beom Han, Woo Jae Kim, Sung-Eui Yoon\textsuperscript{\dag}
\thanks{* equal contribution, \dag corresponding author. 
Supplementary materials are available at \href{https://sigport.org/sites/default/files/docs/SHARE_supplementary.pdf}{\underline{this link}}. }
}
\address{School of Computing, Korea Advanced Institute of Science and Technology (KAIST), Daejeon, Korea}

\begin{document}
\ninept
%


\maketitle
%

\begin{abstract}

While generalizable 3D Gaussian splatting enables efficient, high-quality rendering of unseen scenes, it heavily depends on precise camera poses for accurate geometry. In real-world scenarios, obtaining accurate poses is challenging, leading to noisy pose estimates and geometric misalignments. To address this, we introduce \textbf{SHARE}, a pose-free, feed-forward Gaussian splatting framework that overcomes these ambiguities by joint \textbf{sha}pe and camera \textbf{r}ays \textbf{e}stimation. Instead of relying on explicit 3D transformations, SHARE builds a pose-aware canonical volume representation that seamlessly integrates multi-view information, reducing misalignment caused by inaccurate pose estimates. Additionally, anchor-aligned Gaussian prediction enhances scene reconstruction by refining local geometry around coarse anchors, allowing for more precise Gaussian placement. Extensive experiments on diverse real-world datasets show that our method achieves robust performance in pose-free generalizable Gaussian splatting.

\end{abstract}

\begin{keywords}
Gaussian splatting, 3d-reconstruction, multi-view, generalizable, rendering
\end{keywords}
%

\section{Introduction}

\begin{figure}[!ht]
    \centering
    \includegraphics[width=1.0\linewidth,trim={0 0 16.5cm 0},clip]{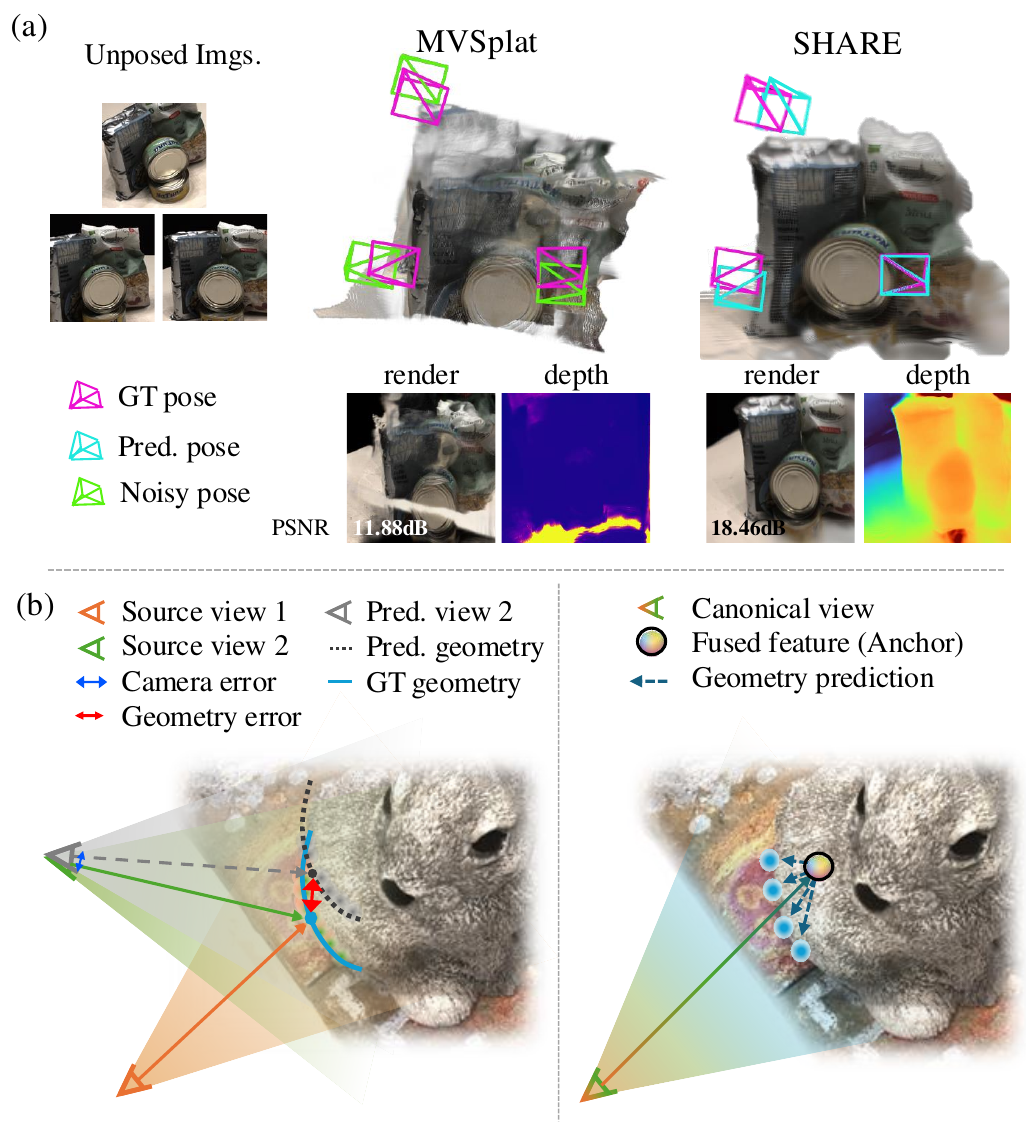}
    \vspace{-0.23in}
    \caption{SHARE predicts geometry, appearance, and relative poses from sparse unposed images. \textbf{(a)} Compared to MVSplat~\cite{mvsplat}, SHARE is more robust to pose noise, producing more accurate geometry and rendering. \textbf{(b)} This is achieved by estimating 3D geometry with an integrated canonical feature instead of aligning geometry predicted from different views.}
    \label{fig:conceptual}
    \vspace{-0.2in}
\end{figure}

Recent advancements in Novel View Synthesis (NVS) and 3D scene reconstruction have been driven by neural implicit representations~\cite{srn,deepsdf,nerf} and explicit volumetric approaches like 3D Gaussian splatting~\cite{3dgs}. Central to this task is the utilization of precise camera poses, which
serve as a fundamental geometric prior coupling the spatial relationship between 3D space and their corresponding 2D-pixel projections across multiple views.
However, the assumption of readily available accurate camera poses is often unsatisfiable in practical scenarios. While Structure-from-Motion (SfM)~\cite{sfm} techniques have long been the go-to solution for obtaining camera poses, they become increasingly unreliable as views become sparser or camera baselines widen. 
Although several previous works~\cite{colmapfreegs, instantsplat,ggrt} propose to reconstruct 3D geometry from sparse views without ground-truth camera poses as input, they often require per-scene optimization~\cite{colmapfreegs,instantsplat} or rely on sequential frames as input~\cite{ggrt}. These limit their general applicability, including diverse scenarios such as large camera baselines or input images with minimal overlaps. 

In a generalizable setting, the pose-free reconstruction becomes particularly challenging as we assume to estimate the camera pose and 3D shape in a network feed-forward manner. When camera pose estimation and shape reconstruction are combined sequentially, even minor inaccuracies in camera pose can lead to significant geometry misalignment, as shown in Fig.~\ref{fig:conceptual} (b, left). This misalignment stems from difficulties in aligning estimated geometry (e.g., depths, point clouds) across different viewpoints by camera transformations and is highly sensitive to errors in estimated relative camera poses.

In this work, we propose a different direction to achieve a robust pose-free 3D reconstruction that circumvents the explicit transformation in the 3D space. Rather than transforming 3D geometries from each viewpoint into one reference camera coordinate (i.e., canonical space) using the predicted camera poses, we first construct a unified cost volume by integrating multi-view features and then estimate 3D Gaussians from that cost volume to represent the entire scene. This approach bypasses the erroneous 3D space transformation and geometry alignment process, making all the multi-view fusion processes in high dimensional latent space, building a fused feature as shown in the right of Fig.~\ref{fig:conceptual} (b).

However, effectively aligning multi-view features from different viewpoints still depends on accurately understanding the relative camera poses of the input images. To achieve this, we jointly estimate relative poses and embed them into the feature alignment process.
Specifically, we use Plücker rays to represent camera poses, where each ray is defined by its direction and moment. Instead of representing the camera as a single rigid transformation relative to a specific reference camera, this formulation encodes a camera pose as a bundle of globally defined rays. It captures view origins and directions pixel-wise in a locally structured way, making it naturally integrable with image features and more robust for multi-view alignment.
By jointly estimating both camera rays and scene geometry within the high-dimensional latent space, our method unifies these two separate problems into a single, consistent optimization process.
The fused features are used to predict pixel-aligned coarse geometry. To capture fine details, we adopt a coarse-to-fine dense Gaussian prediction method, where each predicted point from a pixel serves as an \textit{anchor} in the local 3D region. Then, we predict offset vectors from these anchors to spread to adjacent regions, enhancing the accuracy of the 3D reconstruction.

Experimental results on real-world datasets of different levels of scales, including DTU~\cite{dtu}, BlendedMVS~\cite{blendedmvs}, RealEstate10K~\cite{re10k}, and ACID~\cite{acid} demonstrate that our method achieves robust performance compared to existing generalizable pose-free approaches~\cite{leap, coponerf, smith2023flowcam}. Notably, our approach outperforms these baselines even when compared to pose-dependent Gaussian splatting methods~\cite{pixelsplat,mvsplat} under minimal noise conditions, where the errors in rotation and translation angles are approximately 1 degree. (Fig.~\ref{fig:conceptual} (a))

\section{Related Work}

\noindent
\textbf{Generalizable 3D Gaussian Splatting.}
Generalizable novel view synthesis (NVS) aims to extend applicability by enabling 3D reconstruction without test-time adaptation. Recent generalizable methods in 3D Gaussian splatting~\cite{gpsgaussian, pixelsplat, mvsplat, mvsgaussian, lara} have emerged for fast and efficient reconstruction by predicting dense 3D Gaussians in a feed-forward manner from unseen sparse-view images.
Recent advances include PixelSplat~\cite{pixelsplat}, which leverages probabilistic depth distributions to overcome limitation of optimizing Gaussians, and LatentSplat~\cite{latentsplat}, which incorporates variational auto-encoders (VAEs) for improved depth prediction in challenging scenarios. MVSplat~\cite{mvsplat} and MVSGaussian~\cite{mvsgaussian} enhance geometry fidelity by constructing cost volumes with Multi-View Stereo (MVS). However, these methods rely on precise camera poses, often impractical in real-world applications.

\noindent
\textbf{Pose-Free Generalizable Novel View Synthesis.} 
Recent approaches have explored pose-free solutions for novel view synthesis, aiming to reconstruct scenes without relying on known camera poses.
LEAP~\cite{leap} constrains 3D estimation within a canonical space, eliminating the need for camera poses by employing a shared neural volume for feature similarity-based 2D-3D mapping.  CoPoNeRF~\cite{coponerf} jointly optimizes 6D camera poses and radiance fields through a unified framework integrating 2D correspondence matching, pose estimation, and NeRF rendering. While effective, their NeRF-based backbone results in extended inference time or complicates joint optimization of scene representation and camera poses.

Gaussian splatting-based Pose-free methods~\cite{colmapfreegs,instantsplat} improve optimization efficiency by eliminating the need for camera pose estimation. CF-3DGS~\cite{colmapfreegs} achieves robust reconstruction through sequential frame processing and progressive 3D Gaussian expansion without SfM preprocessing. InstantSplat~\cite{instantsplat} integrates MVS predictions with point-based representations~\cite{dust3r}, optimizing within seconds rather than minutes. However, both methods still require costly test-time adaptation, potentially limiting real-time applications.

Concurrent generalizable approaches~\cite{splatt3r, flash3d} leverage 3D geometric priors from pre-trained foundation models~\cite{unidepth,mast3r} for pose-free reconstruction. However, they often suffer from scale ambiguities, requiring fine-tuning with ground-truth 3D supervision.
While GGRt~\cite{ggrt} only relies on 2D supervision, they employ sequential video frames to estimate poses, where their application is limited to specific input configurations.
In contrast, we propose a 2D-supervised approach that does not rely on explicit 3D priors, offering a more scalable and data-efficient alternative while remaining robust to diverse sparse-view inputs with varying camera baselines.


\section{Methods}
\begin{figure*}[t]
  \centering
  \begin{subfigure}[b]{0.63\textwidth}
   \includegraphics[width=1.0\linewidth, ,trim={0 8.97cm 6.33cm 0},clip]{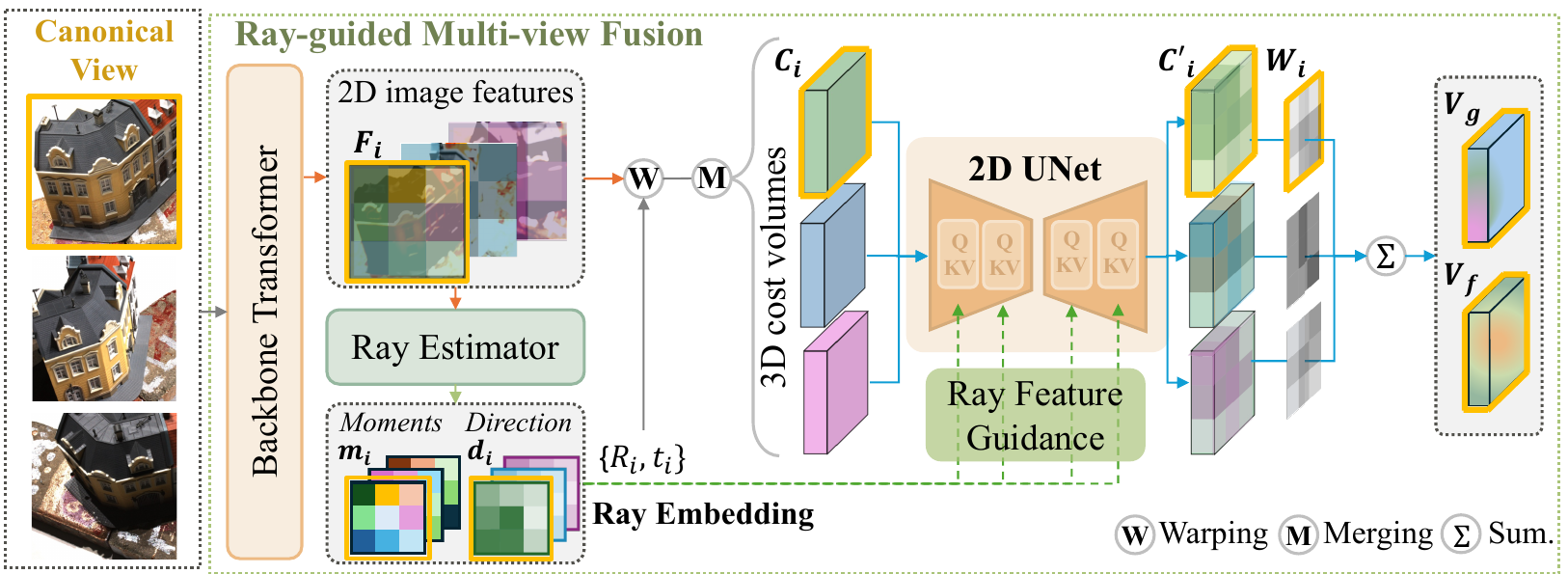}
   \caption{Ray-guided multi-view fusion module}
   \label{fig:ray-guided multi-view fusion}
    \vspace{-1mm}
  \end{subfigure}
   \hfill
  \begin{subfigure}[b]{0.36\textwidth}
   \includegraphics[width=0.9\linewidth,trim={0 10.12cm 20.8cm 0},clip]{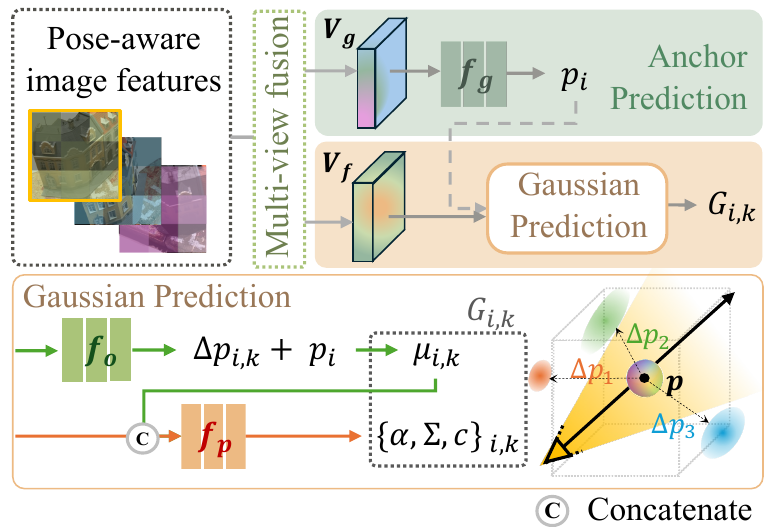}
   \caption{Anchor-aligned dense Gaussian prediction module}
   \label{fig:gaussian prediction}
    \vspace{-1mm}    
  \end{subfigure}
   \caption{\textbf{SHARE Overview.}
    SHARE aims to address geometric misalignment in pose-free 3D Gaussian splatting by jointly estimating relative poses and reconstructing 3D Gaussians in a canonical view. The framework consists of:
    (a) \textbf{Ray-guided multi-view fusion module}: Estimated Plücker rays serve as geometric priors in cost aggregation to construct pose-aware cost volumes, aligning multi-view features in a shared canonical space for improved geometric consistency.
    (b) \textbf{Gaussian prediction module}: Using fused multi-view features, anchor positions from $\textbf{V}_g$ guide the estimation of $k$ Gaussians per region via $\textbf{V}_f$, enabling fine-grained scene reconstruction with reduced misalignment.
    }
   \label{fig:main_method}
    \vspace{-3mm}    
\end{figure*}

Our approach, SHARE, is a pose-free, feed-forward framework for generalizable 3D Gaussian splatting that jointly estimates relative camera poses and reconstructs accurate geometry. Unlike existing methods that fuse the geometry relying on camera transformations in 3D space, SHARE operates transformation only in the latent space, mitigating pose-induced geometry misalignment. As shown in Fig.~\ref{fig:main_method}, our framework consists of two key modules: (1) Ray-guided Multi-view Fusion (Sec.~\ref{sec:ray-guided-fusion} (a)), which constructs a pose-aware canonical volume by embedding Plücker ray-based relative pose information into cost volumes, enabling robust multi-view alignment; and (2) Anchor-aligned Gaussian Prediction (Sec.~\ref{sec:gaussian_prediction} (b)), which refines scene details by predicting dense 3D Gaussians from a unified canonical representation. By integrating pose estimation and 3D reconstruction into a single feed-forward pipeline, our method achieves efficient and scalable pose-free novel view synthesis.

\subsection{Problem Definition}
\label{sec:problem_definition}
Given \( M \) unposed images \( I = \{I_i\}_{i=1}^M \), our goal is to jointly estimate camera poses \(\{\mathbf{R}_i, \mathbf{T}_i\}_{i=1}^M\) and a set of 3D Gaussian primitives \( \{\mathbf{G}_n\}_{n=1}^N \), where each Gaussian \( \textbf{G}_n \) is defined by its position \(\mathbf{\mu}_n\), opacity \( \mathbf{\alpha}_n\), covariance matrix \(\ \mathbf{\Sigma}_n\), and color \(\mathbf{c}_n\). Instead of directly predicting SO(3) camera extrinsic, poses are estimated as a set of Plücker rays. These rays are represented as a combination of direction \(\mathbf{d}\) and momentum \(\mathbf{m}\), denoted by \(\mathbf{P} = \langle \mathbf{d}, \mathbf{m} \rangle\). The conversion from ray representations to camera parameters ($\textbf{R}, \textbf{T}$) is denoted as \( \Psi \), following the formulation in~\cite{raydiffusion}, where the detailed conversion process is provided. The objective mapping function is follows:
\begin{equation}
\Phi_\theta: \{I_i\}_{i=1}^M \mapsto 
\left\{
\begin{aligned}
    &\mathbf{G}_n = \{ \mathbf{\mu}_n, \mathbf{\alpha}_n, \mathbf{\Sigma}_n, \mathbf{c}_n \}_{n=1}^N, \\
    &\{\mathbf{R}_i, \mathbf{T}_i\} = \Psi(\{\mathbf{P}_i^l\}_{l=1}^{P_h \times P_w})
\end{aligned}
\right\},
\end{equation}
where $P_h$ and $P_w$ denote the spatial patch resolution.

\subsection{Ray-Guided Multi-view Fusion}
\label{sec:ray-guided-fusion}
We integrate multi-view features from multiple unposed images of scenes captured from different viewpoints into a fused volume, by selecting one viewpoint as a canonical view and aligning the remaining views to the canonical space. To correctly align these features, we estimate and embed the relative poses as bundles of Plücker rays which provides geometric guidance for the fusion. This ensures pose-aware alignment in the latent space, improving geometric consistency and enabling accurate Gaussian predictions.

\noindent
\textbf{Joint Feature Extraction.}
We employ a matching transformer as our backbone to estimate multi-view features $\textbf{F}_i \in \mathbb{R}^{\frac{H}{4} \times \frac{W}{4} \times C}$. We then estimate patch-wise Plücker rays $ \mathbf{P}_i^l \in \mathbb{R}^{\frac{H}{4} \times \frac{W}{4} \times 6} $ through an additional two-layer lightweight transformer.

\noindent
\textbf{Pose-aware Cost Volumes.}
To effectively integrate multi-view information, it is crucial to establish correspondences across viewpoints while accounting for pose variations.  
Inspired by multi-view Stereo (MVS) techniques~\cite{yao2018mvsnet, ding2022transmvsnet,matchnerf,na2024uforecon,mvsplat}, we construct cost volumes for each viewpoint by extending 2D features into 3D along hypothetical planes~\cite{yao2018mvsnet}, and project them onto other views using the camera poses converted from predicted rays.
Channel-wise correlations are computed between the reference feature $\textbf{F}_i \in \mathbb{R}^{\frac{H}{4} \times \frac{W}{4} \times C}$ and warped features $\{\textbf{F}_{j \rightarrow i}^{d}\}_{d=1}^D \in \mathbb{R}^{\frac{H}{4} \times \frac{W}{4} \times D \times C}$ as follows:
\begin{equation}
\mathbf{C}_i = \frac{\sum_{j \neq i}{\textbf{F}_i \cdot \textbf{F}_{j \rightarrow i}}}{\sqrt{C}} \in \mathbb{R}^{\frac{H}{4} \times \frac{W}{4} \times D}, \quad \forall i \in \{1, 2, \dots, M\},
\end{equation}
where $D$ denotes the number of hypothetical depth candidates. 
After constructing the cost volumes from all viewpoints, we refine them through cost aggregation conditioned on the predicted Plücker rays, using patch-wise cross-attention to embed pose awareness into the cost volumes. Since these rays are locally defined in spatial patches, they can be seamlessly integrated with image features, thus introducing geometric bias in cost volumes.

\noindent
\textbf{Canonical Volume Construction.}
Capturing the same region from different viewpoints often results in varied observations due to view-dependent effects such as occlusion, lighting, and other environmental factors. To address this, we estimate a spatial weight volume $\mathbf{W}_i \in \mathbb{R}^{\frac{H}{4} \times \frac{W}{4}}$ for each cost volume to account for these variations. The refined cost volumes $\mathbf{C'}$ and their corresponding weights $\mathbf{W}$ are then fused into a canonical geometry volume $\mathbf{V}_g \in \mathbb{R}^{\frac{H}{4} \times \frac{W}{4} \times C}$ as follows:
{
\begin{align}
\label{eq:fusion}
\mathbf{V}_g = \sum_{i=1}^{M} \mathbf{W}_i \mathbf{C'}_i 
+ \phi \left( \frac{(\mathbf{C'}_i - \overline{\mathbf{C'}})^2}{M} \oplus \overline{\mathbf{C'}} \right),
\end{align}
}
where $\overline{\mathbf{C}}'$ is the average of all cost volumes, $\oplus$ denotes channel-wise concatenation, and $\phi$ is a CNN layer for channel projection. A mean and variance-based volume is also added following previous MVS techniques~\cite{yao2018mvsnet}. 
This geometry volume $\textbf{V}_g$ is used to estimate the pixel-aligned coarse depths $\textbf{z}$ in the canonical viewpoint.
Similarly, the canonical feature volume $\mathbf{V}_f \in \mathbb{R}^{H \times W \times C}$ is constructed using the same aggregation method with the upsampled multi-view features, which will be used to estimate 3D Gaussian parameters.

\subsection{Anchor-aligned Dense Gaussians Prediction}
\label{sec:gaussian_prediction}
The 3D geometry is estimated as depth values along pixel-aligned rays in the canonical view. 
However, predicted depths capture only the coarse structure of the scene, as they are constrained within a pixel-aligned, ray-bounded space. It might lead to over-smooth geometry and suboptimal rendering, especially in regions with depth discontinuities or complex surfaces. To address this, we introduce \textit{anchor}-aligned coarse-to-fine geometry estimation, where the estimated depth serves as an anchor for the local 3D space, and fine geometry is inferred by predicting local offsets around this geometric center, allowing for more precise surface reconstruction.

Specifically, the initial coarse depth $\textbf{z}$ is computed as the expected value over depth hypothesis planes $\mathcal{D} \in \mathbb{R}^D$, where the probability of each depth candidate is predicted using the feature volume $\mathbf{V}_g$ through a lightweight MLP depth head $f_g$ followed by a softmax function:
\begin{equation}
\mathbf{p}_i = \mathbf{o} + \mathbf{z}_i \mathbf{d}_i, \quad
\mathbf{z} = \sum_{d} \text{softmax}(f_g(\mathbf{V}_g))_d \cdot \mathcal{D}_d,
\end{equation}
where $\mathbf{p}_i$ is the estimated 3D anchor position for pixel $i$, $\mathbf{o}$ is the ray origin, and $\mathbf{d}_i$ is the ray direction.
To enhance spatial fidelity, we estimate $K$ offset vectors for each anchor using the canonical feature volume $\mathbf{V}_{f}$. The offset prediction MLP head $f_o$ outputs $\Delta \mathbf{p}_{i,k} = f_o(\mathbf{V}_{f})$, which is added to the anchor points to obtain the Gaussian means $\{\mu_{i,k}\}_{k=1}^{K}$. The offset vectors are then positionally encoded and another MLP head $f_p$ predicts the remaining Gaussian parameters ($\alpha, \Sigma, c$). Consequently, the offset Gaussians $\mathbf{G}_{i,k} = \{\mu, \alpha, \Sigma, c\}_{i,k}$ for each anchor are defined as:
\begin{equation}
\begin{split}
    \mu_{i,k} &= \mathbf{p}_{i} + \Delta \mathbf{p}_{i,k}, \quad
    \{\alpha, \Sigma, c\}_{i,k} = f_p(\text{PE}(\Delta \mathbf{p}_{i,k}), \mathbf{V}_f),
\end{split}
\end{equation}
where $\Delta \mathbf{p}_{i,k}$ represents the offset for the $k$-th Gaussian relative to anchor $\mathbf{p}_i$, $\text{PE}(\cdot)$ denotes positional encoding, and $\mathbf{G}_{i,k}$ defines the Gaussian parameters. This formulation enables adaptive Gaussian placement, improving spatial detail representation around the anchor points.
Finally, the Gaussians are rasterized into 2D space to render the image.

\subsection{Training and Inference}
SHARE is trained using a combination of photometric and ray regression losses to ensure precise scene reconstruction and pose estimation. The training objectives include minimizing pixel-wise differences through MSE, improving perceptual quality with LPIPS~\cite{lpips}, and optimizing relative camera poses via ray regression. 
In inference time, SHARE only takes unposed RGB images, eliminating the need for ground-truth pose annotations.

\section{Experimental Results}

{\setlength{\tabcolsep}{4pt}
\begin{table}[t]
  \resizebox{\linewidth}{!}{%
  \centering
  \begin{tabular}{p{2cm}l|ccccc}
    \hline
    Method & Pose & \multicolumn{1}{c}{Rot $\downarrow$} & \multicolumn{1}{c}{Trans $\downarrow$} & \multicolumn{1}{c}{PSNR $\uparrow$} & \multicolumn{1}{c}{SSIM $\uparrow$} & \multicolumn{1}{c}{LPIPS $\downarrow$} \\ \midrule
    \multirow{5}{*}{PixelSplat} 
    & \textcolor{gray}{GT} & \textcolor{gray}{--} & \textcolor{gray}{--} & \textcolor{gray}{20.30} & \textcolor{gray}{0.64} & \textcolor{gray}{0.32} \\
    & $\sigma=0.01$ & 1.01 & 1.27 & 16.72 & 0.44 & 0.46 \\
    & $\sigma=0.05$ & 5.06 & 6.30 & 13.39 & 0.32 & 0.63 \\
    & Pred.* & 2.32 & 13.81 & 15.66 & 0.42 & 0.49 \\
    \hline
    \multirow{5}{*}{MVSplat}
    & \textcolor{gray}{GT} & \textcolor{gray}{--} & \textcolor{gray}{--} & \textcolor{gray}{20.41} & \textcolor{gray}{0.66} & \textcolor{gray}{0.26} \\
    & $\sigma=0.01$ & 1.01 & 1.27 & 16.43 & 0.42 & \hlYellowTwo{0.42} \\
    & $\sigma=0.05$ & 5.06 & 6.30 & 12.84 & 0.29 & 0.59 \\
    & Pred.* & 2.32 & 13.81 & 13.96 & 0.34 & 0.55 \\
    \hline
    LEAP & -- & -- & -- & \hlYellowTwo{18.27} & \hlYellowTwo{0.53} & 0.48 \\
    \hline
    Ours & Pred. & 4.79 & 6.06 & \hlYellowOne{\textbf{19.36}} & \hlYellowOne{\textbf{0.61}} & \hlYellowOne{\textbf{0.31}} \\
    \hline
  \end{tabular}%
  }
    \caption{\textbf{Quantitative results on the DTU dataset.} Comparison with generalizable 3DGS and pose-free NeRFs. Pred.* indicates pose estimation using DUSt3R~\cite{dust3r}. \hlYellowOne{\textbf{Best}} and \hlYellowTwo{second-best} results are highlighted. Results with ground-truth poses are shown in gray.}
  \label{tab:sota_dtu}
  \vspace{-2mm}
\end{table}

{\setlength{\tabcolsep}{3.5pt}
\begin{table}[t]
  \resizebox{\linewidth}{!}{%
  \centering
  \begin{tabular}{p{2cm}l|ccccc}
    \hline
    Method & Pose & \multicolumn{1}{c}{Rot $\downarrow$} & \multicolumn{1}{c}{Trans $\downarrow$} & \multicolumn{1}{c}{PSNR $\uparrow$} & \multicolumn{1}{c}{SSIM $\uparrow$} & \multicolumn{1}{c}{LPIPS $\downarrow$} \\ \midrule
    \multirow{5}{*}{PixelSplat} 
    & \textcolor{gray}{GT} & \textcolor{gray}{--} & \textcolor{gray}{--} & \textcolor{gray}{26.08} & \textcolor{gray}{0.86} & \textcolor{gray}{0.14} \\
    & $\sigma=0.01$ & 0.92 & 1.56 & \hlYellowTwo{20.14} & \hlYellowTwo{0.62} & \hlYellowOne{\textbf{0.23}} \\
    & $\sigma=0.05$ & 4.58 & 7.65 & 15.69 & 0.46 & 0.46 \\
    & Pred.* & 1.76 & 12.20 & 11.73 & 0.34 & 0.60 \\
    \hline
    \multirow{5}{*}{MVSplat}  
    & \textcolor{gray}{GT} & \textcolor{gray}{--} & \textcolor{gray}{--} & \textcolor{gray}{26.39} & \textcolor{gray}{0.87} & \textcolor{gray}{0.13} \\
    & $\sigma=0.01$ & 0.92 & 1.56 & 19.99 & \hlYellowTwo{0.62} & \hlYellowOne{\textbf{0.23}} \\
    & $\sigma=0.05$ & 4.58 & 7.64 & 15.12 & 0.44 & 0.45 \\
    & Pred.* & 1.76 & 12.20 & 17.81 & 0.56 & 0.33 \\
    \hline
    FlowCAM & Pred. & 7.43 & 50.66 & 18.24 & 0.60 & 0.64 \\
    CoPoNeRF & Pred. & 3.61 & 12.77 & 19.54 & 0.40 & 0.64 \\
    \hline
    Ours & Pred. &  3.78 &  11.61 & \hlYellowOne{\textbf{21.23}} & \hlYellowOne{\textbf{0.71}} &  \hlYellowTwo{0.26} \\
    \hline
  \end{tabular}%
  }
\caption{\textbf{Quantitative results on the RealEstate10K dataset.} Comparison with generalizable 3DGS and pose-free NeRFs. Pred.* indicates pose estimation using DUSt3R~\cite{dust3r}.}
  \label{tab:sota_re10k}
  \vspace{-0.3cm}
\end{table}

\setlength{\tabcolsep}{3pt}  
\renewcommand{\arraystretch}{1.0}  
\begin{table}[t]
\footnotesize
    \begin{tabular}{lc c@{}c@{}c c@{}c@{}c}
    \toprule
    \multirow{2}{*}{Method} & \multirow{2}{*}[-2pt]{Pose} & \multicolumn{3}{c}{RealEstate10K $\rightarrow$ ACID} & \multicolumn{3}{c}{DTU $\rightarrow$ BlendedMVS} \\
    \cmidrule{3-5} \cmidrule{6-8} 
    & & PSNR$\uparrow$ & SSIM$\uparrow$ & LPIPS$\downarrow$ & PSNR$\uparrow$ & SSIM$\uparrow$ & LPIPS$\downarrow$ \\
    
    \midrule

    \multirow{2}{*}{PixelSplat} & \textcolor{gray}{GT} & \textcolor{gray}{26.84} & \textcolor{gray}{0.81} & \textcolor{gray}{0.18} & \textcolor{gray}{11.64} & \textcolor{gray}{0.20} & \textcolor{gray}{0.67} \\
    & $\sigma=0.01$ & \hlYellowTwo{21.7} & \hlYellowTwo{0.57} & 0.28 & 11.65 & \hlYellowTwo{0.20} & 0.68 \\
        
    \midrule
    
    \multirow{2}{*}{MVSplat} & \textcolor{gray}{GT} & \textcolor{gray}{28.18} & \textcolor{gray}{0.84} & \textcolor{gray}{0.15} & \textcolor{gray}{12.04 }& \textcolor{gray}{0.19} & \textcolor{gray}{0.56} \\
    & $\sigma=0.01$ & 21.65 & \hlYellowTwo{0.57} & \hlYellowTwo{0.27} & \hlYellowTwo{11.92} & \hlYellowTwo{0.20} & \hlYellowOne{\textbf{0.59}} \\
        
    \midrule

    Ours & Pred. & 
    \hlYellowOne{\textbf{23.47}} & 
    \hlYellowOne{\textbf{0.69}} & 
    \hlYellowOne{\textbf{0.26}} & 
    \hlYellowOne{\textbf{12.19}} & \hlYellowOne{\textbf{0.26}} & \hlYellowTwo{0.61} \\
    \bottomrule
    \end{tabular}
    \caption{\textbf{Quantitative comparison of cross-dataset generalization.} Models trained on the RealEstate10K dataset are evaluated on ACID, and models trained on DTU are tested on BlendedMVS. This evaluation highlights the ability to generalize across datasets with varying scene characteristics.}
    \label{tab:cross-dataset}
\end{table}

\noindent
\textbf{Datasets.} 
We train and evaluate our method on DTU~\cite{dtu} and RealEstate10K \cite{re10k}, covering diverse scene scales and camera configurations. 
DTU contains 75 training and 15 testing scenes of small-scale static objects captured from 49 cameras with varying baselines~\cite{na2024uforecon}. 
RealEstate10K includes 67,477 training and 7,289 testing scenes from YouTube real estate videos, representing typical real-world camera movements~\cite{mvsplat}.
We also conduct cross-dataset evaluations on ACID~\cite{acid} and BlendedMVS~\cite{blendedmvs}, trained on RealEstate10K and DTU, respectively.

\noindent
\textbf{Baselines.} 
We compare SHARE against state-of-the-art methods, including generalizable 3D Gaussian splatting (g-3DGS) approaches (PixelSplat~\cite{pixelsplat}, MVSplat~\cite{mvsplat}) and pose-free generalizable NeRFs (LEAP~\cite{leap}, CoPoNeRF~\cite{coponerf}, FlowCAM~\cite{smith2023flowcam}). For g-3DGS methods, we evaluate under predicted poses (DUSt3R~\cite{dust3r}) and corrupted poses with different gaussian noise level (denoted as $\sigma$), following~\cite{sparf}. For NeRF methods, LEAP is evaluated on DTU, while CoPoNeRF and FlowCAM are tested on RealEstate10K. Details of the baseline implementation and comparisons with concurrent pose-free g-3DGS methods are included in the Appendix at this \href{https://sigport.org/sites/default/files/docs/SHARE_supplementary.pdf}{(link)}.

\noindent
\textbf{Metrics.} 
We evaluate novel view synthesis quality using PSNR, SSIM~\cite{ssim}, and LPIPS~\cite{lpips}. Relative pose accuracy is measured with geodesic rotation and angular translation errors, following~\cite{coponerf}.

\noindent
\textbf{Implementation Details.}
We trained our model on DTU with three context and one target image, and on RealEstate10K with two context and three target images, using a resolution of 256$\times$256. The multi-view fusion backbone uses six matching Transformer layers~\cite{mvsplat}, and dense ray prediction employs a modified 2-layer Transformer~\cite{raydiffusion}. The number of predicted anchors matches image resolution, with three Gaussian primitives per anchor. Training ran for 140K iterations on DTU and 300K on RealEstate10K using the Adam optimizer on an NVIDIA RTX 4090 GPU.

\vspace{-0.1cm}
\subsection{Novel View Synthesis.}
On the DTU dataset (Table~\ref{tab:sota_dtu}), our method outperforms g-3DGS baselines under noisy pose conditions and achieves LPIPS~\cite{lpips} and SSIM~\cite{ssim} scores comparable to those obtained with ground-truth poses. This demonstrates that our pose-aware multi-view fusion effectively mitigates the impact of pose uncertainty, ensuring robust reconstruction performance. Notably, g-3DGS methods are highly sensitive to pose inaccuracies, with even minor Gaussian noise (\(\sigma=0.01\)) causing significant degradation, especially in sparse-view settings.
On the RealEstate10K dataset (Table~\ref{tab:sota_re10k}), our approach consistently outperforms both g-3DGS with estimated poses and pose-free NeRF methods~\cite{coponerf,smith2023flowcam}, demonstrating its scalability to complex, large-scale scenes.
Figure~\ref{fig:result_sota} shows qualitative comparisons, where our method achieves superior fidelity and geometric consistency, further validating its effectiveness in challenging pose-free scenarios.

\noindent
\textbf{Cross-dataset Generalization.}
Table~\ref{tab:cross-dataset} presents the cross-dataset generalization results, where SHARE, trained on DTU~\cite{dtu}, is evaluated on BlendedMVS~\cite{blendedmvs}, and trained on RealEstate10K~\cite{re10k}, is tested on ACID~\cite{acid}. SHARE achieves robust performance across all metrics, showing generalizability across datasets. Additional qualitative results and dataset details are provided in the Appendix~\href{https://sigport.org/sites/default/files/docs/SHARE_supplementary.pdf}{(link)}.

{\setlength{\tabcolsep}{7pt}
\begin{table}[t]
  \resizebox{\linewidth}{!}{%
  \centering
  \begin{tabular}{p{4.0cm}|ccc}
    \hline
    Method & PSNR $\uparrow$ & SSIM $\uparrow$ & LPIPS $\downarrow$ \\ \hline
    \textit{w/o pose embedding} & 17.52 & 0.54 & 0.37 \\
    \textit{anchor only} & 14.05 & 0.38 & 0.57 \\
    \textit{w/ 1 offsets} & 19.11 & 0.60 & 0.32 \\
    \textit{w/ 2 offsets} & \hlYellowTwo{19.33} & \hlYellowTwo{0.61} & \hlYellowOne{\textbf{0.30}} \\
    \hline
    Ours & \hlYellowOne{\textbf{19.36}} & \hlYellowOne{\textbf{0.61}} & \hlYellowTwo{0.31} \\
    \hline
  \end{tabular}%
  }
    \caption{\textbf{Ablation study on the DTU dataset.} Evaluation of pose embedding and offsets on rendering quality. The full model achieves the best results, with pose embedding and increased offsets significantly improving performance.}
  \label{tab:ablation_dtu}
  \vspace{-0.4cm}
\end{table}

\begin{figure*}[th!]
    \centering
    \includegraphics[width=0.85\linewidth,trim={0 19cm 7cm 0cm},clip]{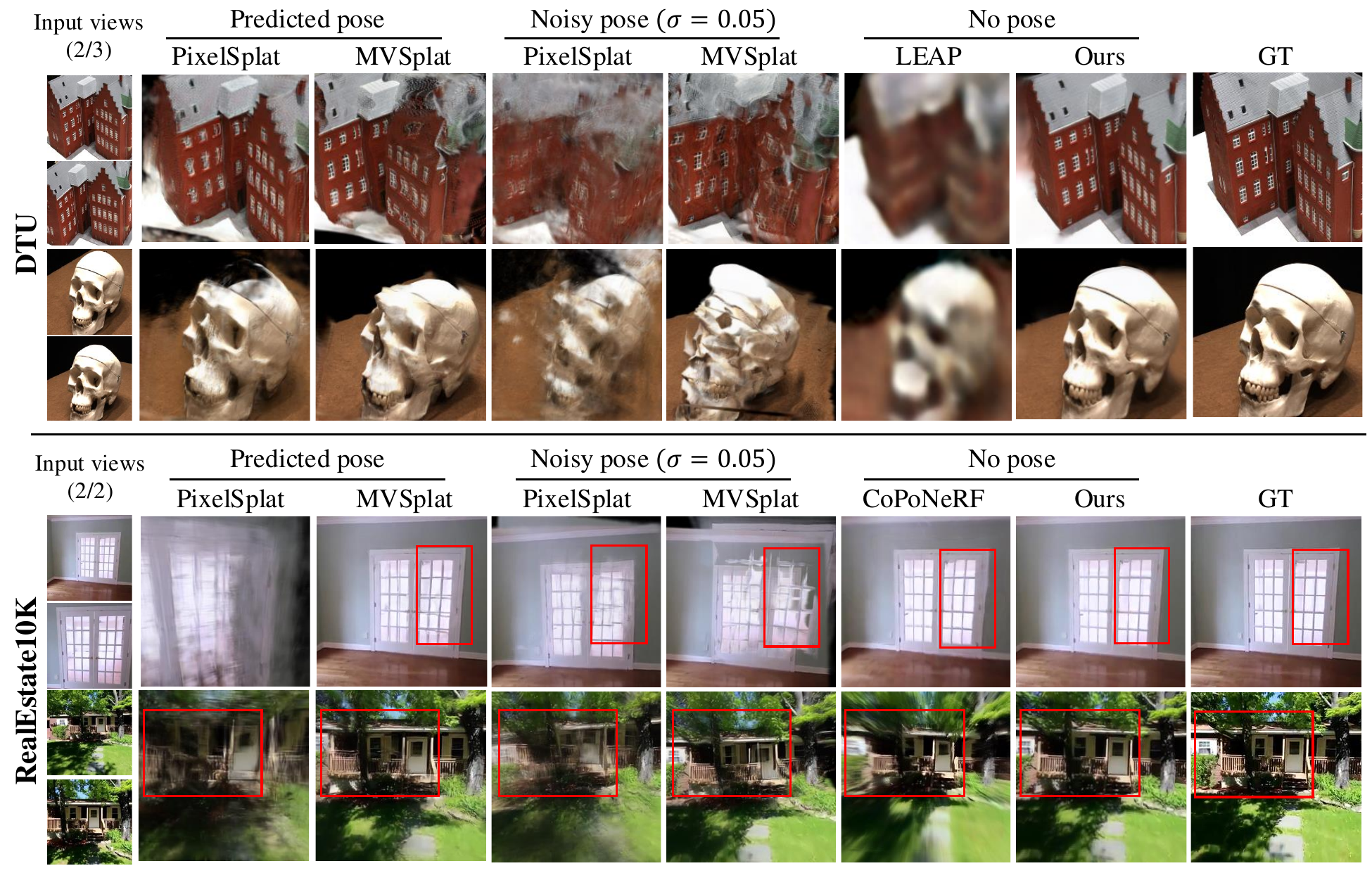}
    \caption{\textbf{Qualitative results on DTU and RealEstate10K datasets.} We visualized rendering results of multiple scenes from DTU and RealEstate10K datasets. Our method captures fine details with correct geometry.}
    \label{fig:result_sota}
    \vspace{-0.1in}
\end{figure*}

\vspace{-0.1cm}
\subsection{Ablation Study}

\begin{figure}[t]
    \centering
    \begin{minipage}{0.4\linewidth}
        \centering
        \includegraphics[width=\linewidth,trim={0 7.5cm 23.2cm 0},clip]{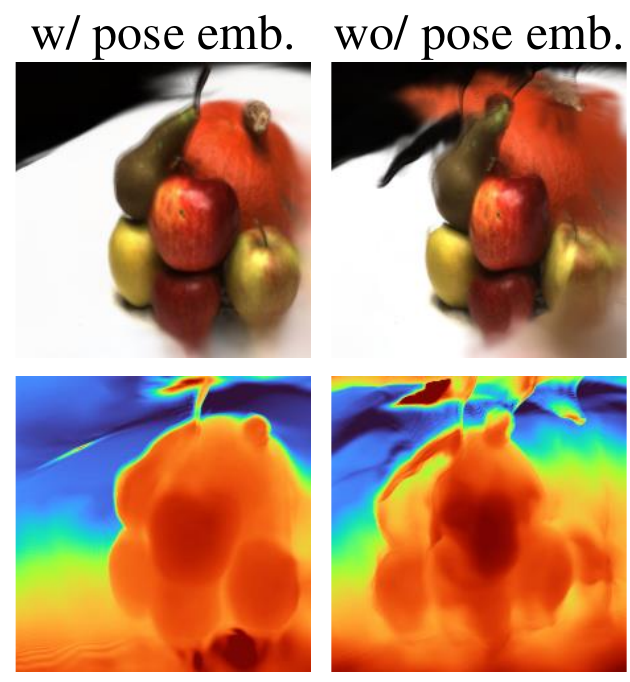}
        \caption{\small\textbf{Effect of pose embedding on DTU.} 
        }
        \label{fig:ablation_pemb}
    \end{minipage}
    \hfill
    \begin{minipage}{0.56\linewidth}
        \centering
        \includegraphics[width=\linewidth,trim={0 7cm 18.3cm 0},clip]{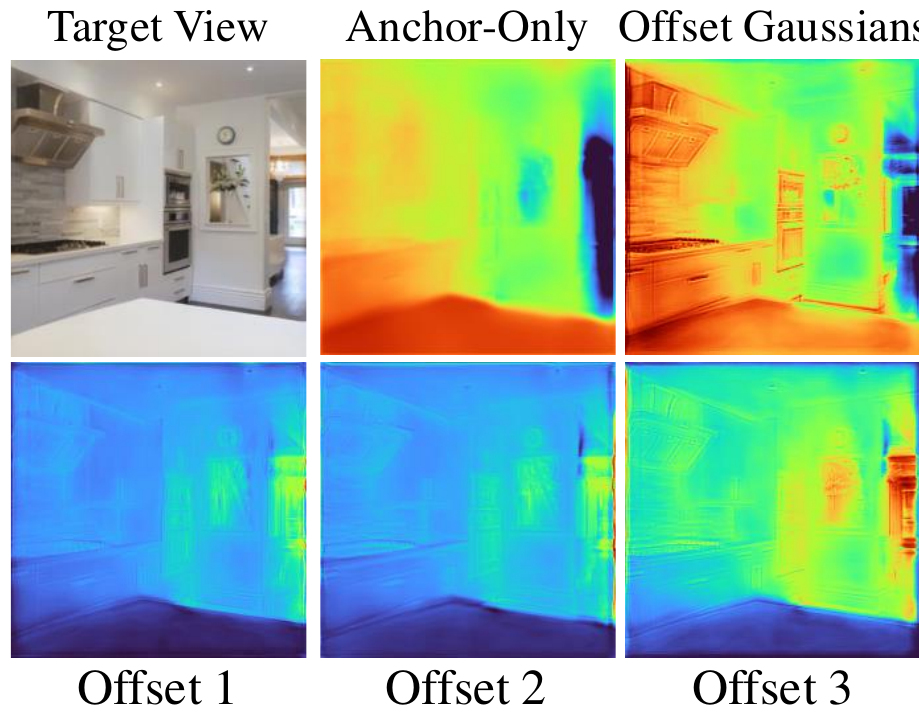}
        \caption{\small\textbf{Anchor-aligned Gaussian prediction on RealEstate10K.} 
        }
        \label{fig:ablation_offset}
    \end{minipage}
    \vspace{-0.5cm}
\end{figure}

\noindent
\textbf{Pose Embedding.}
Integrating per-pixel pose priors enhances geometric consistency by improving spatial awareness in 3D estimation. As shown in Table~\ref{tab:ablation_dtu} and Fig.~\ref{fig:ablation_pemb}, the absence of pose embedding leads to severe geometric degradation in novel view synthesis. The improved depth and appearance when incorporating pose embedding indicate that pose-aware multi-view fusion effectively captures geometric details and establishes stronger multi-view relationships.

\noindent
\textbf{Anchor-Aligned Gaussians Prediction.}
We evaluate anchor-based Gaussian prediction by examining how offset predictions capture geometric details around anchor points. Table~\ref{tab:ablation_dtu} shows that increasing offset density improves rendering quality, confirming finer detail coverage. Visualizing $K=3$ offsets further reveals that each corresponds to distinct depth regions (Fig.~\ref{fig:ablation_offset}), indicating that our learned representation effectively supports geometry estimation across viewpoints.
\vspace{-0.5cm}
\section{Conclusion}  
In this paper, we introduce SHARE, a pose-free framework for generalizable 3D Gaussian splatting. By leveraging ray-guided multi-view fusion and anchor-aligned Gaussian prediction, SHARE effectively mitigates geometric misalignment caused by pose uncertainty.  
Extensive evaluations on diverse datasets demonstrate that SHARE outperforms existing generalizable pose-free reconstruction methods with improved geometric consistency. Furthermore, our approach generalizes well across datasets without requiring test-time pose optimization, making it an efficient and scalable solution for pose-free novel view synthesis and 3D reconstruction.  

\noindent
\textbf{Limitations.}
While SHARE demonstrates strong pose-free generalization, its performance can be limited in scenes with minimal view overlap, such as sparse 360-degree inputs. These aspects present opportunities for future work to further enhance its applicability across broader scenarios.

\vspace{-0.1cm}
\section{acknowledgements}
%
This work was supported by the Institute of Information \& Communications Technology Planning \& Evaluation (IITP) grant (No. RS-2023-00237965, Recognition, Action and Interaction Algorithms for Open-world Robot Service), IITP–ITRC (Information Technology Research Center) support program (No. IITP-2025-RS-2020-II201460), and the National Research Foundation of Korea (NRF) grant (No. RS-2023-00208506), all funded by the Korea government (MSIT).



\vspace{-0.1cm}
\begin{thebibliography}{10}
    \bibitem{mvsplat}
    Y.~Chen, H.~Xu, C.~Zheng, B.~Zhuang, M.~Pollefeys, A.~Geiger, T.~J. Cham, and J.~Cai,
    \newblock ``MVSplat: Efficient 3D Gaussian splatting from sparse multi-view images,''
    \newblock {\em Proc. ECCV}, pp. 370--386, 2024.
    
    \bibitem{srn}
    V.~Sitzmann, M.~Zollh{\"o}fer, and G.~Wetzstein,
    \newblock ``Scene representation networks: Continuous 3D-structure-aware neural scene representations,''
    \newblock {\em Advances in Neural Information Processing Systems}, vol. 32, 2019.
    
    \bibitem{deepsdf}
    J.~J. Park, P.~Florence, J.~Straub, R.~Newcombe, and S.~Lovegrove,
    \newblock ``DeepSDF: Learning continuous signed distance functions for shape representation,''
    \newblock {\em Proc. CVPR}, pp. 165--174, 2019.
    
    \bibitem{nerf}
    B.~Mildenhall, P.~P. Srinivasan, M.~Tancik, J.~T. Barron, R.~Ramamoorthi, and R.~Ng,
    \newblock ``NeRF: Representing scenes as neural radiance fields for view synthesis,''
    \newblock {\em Commun. ACM}, vol. 65, no. 1, pp. 99--106, 2021.
    
    \bibitem{3dgs}
    B.~Kerbl, G.~Kopanas, T.~Leimk{\"u}hler, and G.~Drettakis,
    \newblock ``3D Gaussian splatting for real-time radiance field rendering,''
    \newblock {\em ACM Trans. Graph.}, vol. 42, no. 4, pp. 139--1, 2023.
    
    \bibitem{sfm}
    N.~Snavely, S.~M. Seitz, and R.~Szeliski,
    \newblock ``Photo tourism: exploring photo collections in 3D,''
    \newblock {\em ACM SIGGRAPH 2006 Papers}, pp. 835--846, 2006.
    
    \bibitem{colmapfreegs}
    Y.~Fu, S.~Liu, A.~Kulkarni, J.~Kautz, A.~A. Efros, and X.~Wang,
    \newblock ``COLMAP-Free 3D Gaussian Splatting,''
    \newblock {\em arXiv preprint arXiv:2312.07504}, 2023.
    
    \bibitem{instantsplat}
    Z.~Fan, W.~Cong, K.~Wen, K.~Wang, J.~Zhang, X.~Ding, D.~Xu, B.~Ivanovic, M.~Pavone, G.~Pavlakos, et~al.,
    \newblock ``InstantSplat: Unbounded sparse-view pose-free Gaussian splatting in 40 seconds,''
    \newblock {\em arXiv preprint arXiv:2403.20309}, 2024.
    
    \bibitem{ggrt}
    H.~Li, Y.~Gao, D.~Zhang, C.~Wu, Y.~Dai, C.~Zhao, H.~Feng, E.~Ding, J.~Wang, and J.~Han,
    \newblock ``GGRt: Towards generalizable 3D Gaussians without pose priors in real-time,''
    \newblock {\em arXiv preprint arXiv:2403.10147}, 2024.
    
    \bibitem{dtu}
    R.~Jensen, A.~Dahl, G.~Vogiatzis, E.~Tola, and H.~Aan{\ae}s,
    \newblock ``Large scale multi-view stereopsis evaluation,''
    \newblock {\em Proc. CVPR}, pp. 406--413, 2014.
    
    \bibitem{re10k}
    T.~Zhou, R.~Tucker, J.~Flynn, G.~Fyffe, and N.~Snavely,
    \newblock ``Stereo magnification: Learning view synthesis using multiplane images,''
    \newblock {\em ACM Trans. Graph.}, vol. 37, 2018.
    
    \bibitem{leap}
    H.~Jiang, Z.~Jiang, Y.~Zhao, and Q.~Huang,
    \newblock ``LEAP: Liberate sparse-view 3D modeling from camera poses,''
    \newblock {\em arXiv preprint arXiv:2310.01410}, 2023.
    
    \bibitem{coponerf}
    S.~Hong, J.~Jung, H.~Shin, J.~Yang, S.~Kim, and C.~Luo,
    \newblock ``Unifying correspondence pose and NeRF for generalized pose-free novel view synthesis,''
    \newblock {\em Proc. CVPR}, pp. 20196--20206, 2024.
    
    \bibitem{smith2023flowcam}
    C.~Smith, Y.~Du, A.~Tewari, and V.~Sitzmann,
    \newblock ``FlowCAM: Training generalizable 3D radiance fields without camera poses via pixel-aligned scene flow,''
    \newblock {\em arXiv preprint arXiv:2306.00180}, 2023.
    
    \bibitem{pixelsplat}
    D.~Charatan, S.~L. Li, A.~Tagliasacchi, and V.~Sitzmann,
    \newblock ``PixelSplat: 3D Gaussian splats from image pairs for scalable generalizable 3D reconstruction,''
    \newblock {\em Proc. CVPR}, pp. 19457--19467, 2024.
    
    \bibitem{gpsgaussian}
    S.~Zheng, B.~Zhou, R.~Shao, B.~Liu, S.~Zhang, L.~Nie, and Y.~Liu,
    \newblock ``GPS-Gaussian: Generalizable pixel-wise 3D Gaussian splatting for real-time human novel view synthesis,''
    \newblock {\em Proc. CVPR}, 2024.
    
    \bibitem{mvsgaussian}
    T.~Liu, G.~Wang, S.~Hu, L.~Shen, X.~Ye, Y.~Zang, Z.~Cao, W.~Li, and Z.~Liu,
    \newblock ``MVSGaussian: Fast generalizable Gaussian splatting reconstruction from multi-view stereo,''
    \newblock {\em arXiv preprint arXiv:2405.12218}, 2024.
    
    \bibitem{latentsplat}
    C.~Wewer, K.~Raj, E.~Ilg, B.~Schiele, and J.~E. Lenssen,
    \newblock ``LatentSplat: Autoencoding variational Gaussians for fast generalizable 3D reconstruction,''
    \newblock {\em Proc. ECCV}, pp. 456--473, 2024.
    
    
    \bibitem{flash3d}
    S.~Szymanowicz, E.~Insafutdinov, C.~Zheng, D.~Campbell, J.~F. Henriques, C.~Rupprecht, and A.~Vedaldi,
    \newblock ``Flash3D: Feed-forward generalizable 3D scene reconstruction from a single image,''
    \newblock {\em arXiv preprint arXiv:2406.04343}, 2024.
    
    \bibitem{splatt3r}
    B.~Smart, C.~Zheng, I.~Laina, and V.~A. Prisacariu,
    \newblock ``Splatt3R: Zero-shot Gaussian splatting from uncalibrated image pairs,''
    \newblock {\em arXiv preprint arXiv:2408.13912}, 2024.
    
    \bibitem{raydiffusion}
    J.~Y. Zhang, A.~Lin, M.~Kumar, T.~H.~Yang, D.~Ramanan, and S.~Tulsiani,
    \newblock ``Cameras as rays: Pose estimation via ray diffusion,''
    \newblock {\em Proc. ICLR}, 2024.
    
    \bibitem{ding2022transmvsnet}
    Y.~Ding, W.~Yuan, Q.~Zhu, H.~Zhang, X.~Liu, Y.~Wang, and X.~Liu,
    \newblock ``TransMVSNet: Global context-aware multi-view stereo network with transformers,''
    \newblock {\em Proc. CVPR}, pp. 8585--8594, 2022.
    
    \bibitem{matchnerf}
    Y.~Chen, H.~Xu, Q.~Wu, C.~Zheng, T.~J. Cham, and J.~Cai,
    \newblock ``Explicit correspondence matching for generalizable neural radiance fields,''
    \newblock {\em arXiv preprint arXiv:2304.12294}, 2023.
    
    \bibitem{na2024uforecon}
    Y.~Na, W.~J. Kim, K.~B. Han, S.~Ha, and S.-E. Yoon,
    \newblock ``UFORecon: Generalizable sparse-view surface reconstruction from arbitrary and unfavorable sets,''
    \newblock {\em Proc. CVPR}, pp. 5094--5104, 2024.
    
    \bibitem{yao2018mvsnet}
    Y.~Yao, Z.~Luo, S.~Li, T.~Fang, and L.~Quan,
    \newblock ``MVSNet: Depth inference for unstructured multi-view stereo,''
    \newblock {\em Proc. ECCV}, pp. 767--783, 2018.
    
    \bibitem{lara}
    A.~Chen, H.~Xu, S.~Esposito, S.~Tang, and A.~Geiger,
    \newblock ``LaRa: Efficient large-baseline radiance fields,''
    \newblock {\em Proc. ECCV}, 2024.
    
    \bibitem{lpips}
    R.~Zhang, P.~Isola, A.~A. Efros, E.~Shechtman, and O.~Wang,
    \newblock ``The unreasonable effectiveness of deep features as a perceptual metric,''
    \newblock {\em Proc. CVPR}, pp. 586--595, 2018.
    
    \bibitem{acid}
    A.~Liu, R.~Tucker, V.~Jampani, A.~Makadia, N.~Snavely, and A.~Kanazawa,
    \newblock ``Infinite nature: Perpetual view generation of natural scenes from a single image,''
    \newblock {\em Proc. ICCV}, pp. 14458--14467, 2021.
    
    \bibitem{blendedmvs}
    Y.~Yao, Z.~Luo, S.~Li, J.~Zhang, Y.~Ren, L.~Zhou, T.~Fang, and L.~Quan,
    \newblock ``BlendedMVS: A large-scale dataset for generalized multi-view stereo networks,''
    \newblock {\em Proc. CVPR}, pp. 1790--1799, 2020.
    
    \bibitem{dust3r}
    S.~Wang, V.~Leroy, Y.~Cabon, B.~Chidlovskii, and J.~Revaud,
    \newblock ``Dust3R: Geometric 3D vision made easy,''
    \newblock {\em Proc. CVPR}, pp. 20697--20709, 2024.
    
    \bibitem{sparf}
    P.~Truong, M.-J. Rakotosaona, F.~Manhardt, and F.~Tombari,
    \newblock ``Sparf: Neural radiance fields from sparse and noisy poses,''
    \newblock {\em Proc. CVPR}, pp. 4190--4200, 2023.
    
    \bibitem{ssim}
    Z.~Wang, A.~C.~Bovik, H.~R.~Sheikh, and E.~P.~Simoncelli,
    \newblock ``Image quality assessment: From error visibility to structural similarity,''
    \newblock {\em IEEE Trans. Image Process.}, vol. 13, no. 4, pp. 600--612, 2004.
    
    \bibitem{cat3d}
    R.~Gao, A.~Holynski, P.~Henzler, A.~Brussee, R.~Martin-Brualla, P.~Srinivasan, J.~T. Barron, and B.~Poole,
    \newblock ``Cat3D: Create anything in 3D with multi-view diffusion models,''
    \newblock {\em arXiv preprint arXiv:2405.10314}, 2024.
    
    \bibitem{lgm}
    J.~Tang, Z.~Chen, X.~Chen, T.~Wang, G.~Zeng, and Z.~Liu,
    \newblock ``LGM: Large multi-view Gaussian model for high-resolution 3D content creation,''
    \newblock {\em Proc. ECCV}, pp. 1--18, 2024.

    \bibitem{unidepth}
    L.~Piccinelli, Y.-H.~Yang, C.~Sakaridis, M.~Segu, S.~Li, L.~Van Gool, and F.~Yu,
    \newblock ``UniDepth: Universal monocular metric depth estimation,''
    \newblock {\em Proc. CVPR}, pp. 10106--10116, 2024.

    \bibitem{mast3r}
    V.~Leroy, Y.~Cabon, and J.~Revaud,
    \newblock ``Grounding image matching in 3D with MAST3R,''
    \newblock {\em Proc. ECCV}, pp. 71--91, 2024.



\end{thebibliography}
\twocolumn

\end{document}